\pdfoutput=1
\documentclass[11pt]{article}

\usepackage[]{acl}

\usepackage{times}
\usepackage{latexsym}

\usepackage[T1]{fontenc}
\usepackage[utf8]{inputenc}

\usepackage{microtype}

\usepackage{amsmath, amsfonts}
\usepackage{graphicx}
\usepackage{subcaption}
\usepackage{url, hyperref}
\usepackage{xcolor}

\title{A Cross-Domain Evaluation of Approaches for Causal Knowledge Extraction}
\author{Anik Saha \\ Rensselaer Polytechnic Inst. \\ {\small \texttt{sahaa@rpi.edu}} \\ \And
  Oktie Hassanzadeh \\ IBM Research \\ {\small \texttt{hassanzadeh@us.ibm.com}}  \And 
  Alex Gittens \\ Rensselaer Polytechnic Inst. \\ {\small \texttt{gittea@rpi.edu}} \AND
  Jian Ni \\ IBM Research \\ {\small \texttt{nij@us.ibm.com}} \And
  Kavitha Srinivas \\ IBM Research \\ {\small \texttt{kavitha.srinivas@ibm.com}} \And 
  B{\"u}lent Yener \\ Rensselaer Polytechnic Inst. \\ {\small \texttt{yener@cs.rpi.edu}}
  }

\begin{document}
\maketitle

\begin{abstract}
Causal knowledge extraction is the task of extracting relevant causes and effects from text by detecting the causal relation. 
Although this task is important for language understanding and knowledge discovery, recent works in this domain have largely focused on binary classification of a text segment as causal or non-causal.
In this regard, we perform a thorough analysis of three sequence tagging models for causal knowledge extraction and compare it with a span based approach to causality extraction.
Our experiments show that embeddings from pre-trained language models (e.g. BERT) provide a significant performance boost on this task compared to previous state-of-the-art models with complex architectures. 
We observe that span based models perform better than simple sequence tagging models based on BERT across all 4 data sets from diverse domains with different types of cause-effect phrases.
% For our experiments, we convert data sets from different domains to a standard format for a comprehensive evaluation of different models on this task.
% We analyze the variability in the data sets and apply different evaluation methods to understand the important factors in designing neural models for causal knowledge extraction.
\end{abstract}

\section{Introduction}
% \textcolor{red}{Main contribution of the paper is to compare four recent state-of-the-art models for cause and effect extraction across multiple domains, to elucidate the importance of: (i) differences in the data sets/domains, (ii) architectures of the models, and (iii) evaluation metrics towards their performance. Because each paper introduced and evaluated model within a specific domain (correct claim?), need a direct comparison of models across domains to understand these factors. Introduce a span-based causal knowledge extraction baseline, and show it consistently outperforms these baselines in terms of the most demanding accuracy metric when maximum span length is tuned well for the domain. Large variance in performance of models across domains point towards the need to explicitly evaluate each model across domains when they are introduced, as well as the need to design models explicitly to work well across domains.}

Automatic extraction of causal knowledge is an important task for language understanding.
The causal relationship between entities and events can be used to build a forecasting model for future events.
Causality information is expressed in many different forms in natural language.
There might be explicit indicators like \textit{because}, \textit{therefore}, \textit{result in} etc.
A sentence can also encode causal relationship without using any causal connective word.
So rule based methods are limited in their ability to extract causal knowledge from such texts.
For a causality extraction system, it is important to understand the semantics of the text and recognize the pattern of sentences containing causal knowledge.
The flexibility of neural models to represent any pattern in data makes them a strong candidate for this task.

To extract causal relationship in a text segment, the first step is to determine the existence of causal relationship. 
The next step is extracting the related cause and effect entities.
Figure \ref{fig:example} shows some examples of causes and effects in the data sets used in experiments for this paper. 
In recent years, there has been an increasing interest in developing neural models for causality extraction. 
These models focus on solving the binary problem of detecting if a sequence of words contains causal information.
But we are more interested in extracting the related cause and effect pair from a causal text segment. 
Surprisingly, there is a scarcity of studies on performing this important knowledge extraction task both in terms of models and data sets.
In this paper, we assemble data sets with cause and effect span labels from different domains and convert them to a standard format \cite{hosseini2021predicting}.
% \footnote{\url{https://github.com/phosseini/CREST}}
Then we evaluate the performance of four neural models and analyse the results to determine the effect of the model classes and different data set attributes.
We also study two modes of evaluation for the span extraction or sequence labeling tasks.

\begin{figure}
    \centering
    \begin{subfigure}{\linewidth}
        \begin{align*}
            &\text{The drug lords who claimed responsibility said} \\
            &\underbrace{\text{\color{blue} they would blow up the Bogota newspaper's}}_{\text{effect}} \\
            &\underbrace{\text{\color{blue} offices}}_{\text{effect}}
            \text{ if}
            \underbrace{\text{\color{red} it continued to distribute in Medellin.}}_{\text{cause}}
        \end{align*}
        \caption{BeCauSE data set}
    \end{subfigure}
    \begin{subfigure}{\linewidth}
        \begin{align*}
            \underbrace{\text{\color{blue} The light}}_{\text{effect}}
            \text{ in the background is from}
            \underbrace{\text{\color{red} the sunrise}}_{\text{cause}}
        \end{align*}
        \caption{SemEval-2010 data set}
    \end{subfigure}
    \caption{Example Causal Sentence}
    \label{fig:example}
\end{figure}

The main contribution of the paper is to compare three recent state-of-the-art models for cause and effect extraction across multiple domains, to elucidate the importance of: (i) differences in the data sets/domains, (ii) architectures of the models, and (iii) evaluation metrics towards their performance.
Because each paper introduced and evaluated model within a specific domain, we need a direct comparison of models across domains to understand these factors. 
We apply a span-based relation extraction model to the causal knowledge extraction task, and show it consistently outperforms these baselines in terms of the most demanding accuracy metric when maximum span length is tuned well for the domain. 
Large variance in performance of models across domains point towards the need to explicitly evaluate each model across domains when they are introduced, as well as the need to design models explicitly to work well across domains.

% If we can train a model to label the spans of words in a sentence as cause or effect, it will accomplish both tasks.
% We build our model on the pre-trained lanugage model BERT as it has become the de facto in different lanugage understanding tasks.

\section{Related Work}
Research on causal relation extraction have been mostly focused on detecting the existence of causality. 
Prior works can be broadly categorized into 3 classes based on their primary objective; text classification, causal relation identification and cause-effect extraction / sequence labelling \citep{xu2020review}.

\paragraph{Text Classification} In text classification, the task is to determine if a text segment contains causal information. 
The model is not required to extract the cause or effect entities in the text.
\citet{blanco2008causal} manually selected features to train a bagging tree for classifying sentences as causal or not causal. 
The features were based on the connector words, their modifiers, the class and tense of the cause and effect verbs.
\citet{hidey-mckeown-2016-identifying} trained a linear SVM model using features created from parallel corpus and lexical features from WordNet, FrameNet, VerbNet.
They also created a data set for causal sentence classification using the parallel articles from Simple and English Wikipedia.

\paragraph{Causal Relation Identification} Here, the task is to judge whether a pair of entities or events have a causal relationship given a context, i.e. a sentence.
This is also a classification problem approached by several recent papers using deep learning methods.
\citet{kruengkrai2017improving} trained a multi-column CNN model with 8 columns to determine if a causality relationship exist between two event mentions.
They used 5 columns to feed vector representation of the event mentions and different fragments of the context. 
The other 3 columns take in background knowledge from different sources.
\citet{zhao2016event} proposed a restricted hidden naive bayes (RHNB) model to extraction candidate phrases from sentences and determine their relationship. 
\citet{li2019knowledge} used background knowledge from WordNet and FrameNet to train a knowledge-oriented CNN model for causality extraction.
There is a knowledge-oriented channel and a data-oriented channel in the model to process lexical knowledge and the input sentence respectively.
\citet{liu-etal-2020-knowledge} addressed the event causality identification using the pre-trained language model, BERT as an encoder. 
They extracted knowledge from ConceptNet to augment the input sentence to the model. 
A masking loss was also used to predict the event mentions in text. 

\paragraph{Cause-effect Extraction} In this task the goal is to find the span of the cause and effect entities or events in a text segment if there exists a causal relationship.
We are interested in this task as it helps us utilize the vast amount of text available online to `discover' causal relationships. 
\citet{dasgupta2018automatic} developed a liguistically informed Bidirectional LSTM model by annotating the words in the SemEval-2010 data set with cause, effect and causal connective labels. 
They evaluated their performance on cause, effect and causal connective word extraction separately. 
\citet{li2021causality} also separately annotated the SemEval-2010 data set and trained a self-attentive BiLSTM-CRF model with Flair word embeddings. 
They also addressed the problem of multiple causality where one cause can be related to more than one effect.
\citet{moghimifar2020domain} developed a graph convolution network (GCN) added to the output of BiLSTM using the dependency relationship between words.
They created a data set by labeling medical articles from Wikipedia and proposed a domain adaptation technique using adversarial training method.

\section{Model Description}

We train 4 different models on labeled causal relation data sets. 
They are broadly divided into two classes: sequence tagging model and span based model.

\subsection{Sequence Tagging Models}
Here, the cause-effect extraction task is modeled as a sequence tagging task. 
The model predicts a tag for each word in the input sentence or paragraph.
We use tagging schemes like Named Entity Recognition (NER) to train these models.

\subsubsection{SCITE}
\paragraph{Flair-BiLSTM-CRF} This is a bidirectional LSTM model with a CRF output layer from \cite{li2021causality}.
They use pre-trained Flair \citep{akbik2018coling} embeddings and pre-trained word embeddings from \cite{komninos2016dependency}.
A character level convolutional network is used to build word embeddings from character embeddings.
The embedding for each word is a concatentation of these 3 types of embeddings. 
\[
E_w = [E_c; E_F; E_d]
\]
where $E_c$ is the word embedding from character level embeddings, 
$E_F$ is the pre-trained Flair embedding and 
$E_d$ is the dependency based word embedding trained in \cite{komninos2016dependency}.

The bidirectional LSTM has a forward and a backward LSTM layer that takes in the word embeddings $E_w$. 
The output embeddings from forward and backward LSTM layers are concateneted to obtain the output embeddings.
Next, they add a multi-headed self-attention layer over these embeddings similar to a transformer model. 
Finally, there is a CRF layer to model the sequential relation of words in representing causality.
A BIO tagging scheme is used to label the cause and effect spans in the data set.
The model is trained to maximize the log likelihood of the correct label sequence.

\subsubsection{GCE}
\paragraph{BiLSTM-GCN} We use the graph convolution encoder (GCE) model from \cite{moghimifar2020domain}.
Here, the graph for a sentence is built using dependency parsing relations. 
In the adjacency matrix of the graph, the corresponding entry of two tokens related according to dependency parse has a value of 1 and all other values are zero. 
A graph convolution network uses the graph structure based on this adjacency matrix to obtain embeddings for words (nodes). 
The input to the GCN is a sequence of embeddings from a \text{bidi}rectional LSTM built on top ofFFNN( word embeddings. 
\[
\mathbf{h}_{\text{GCN}}(X) = f^{\text{pool}} (\text{GCN} (\text{BiLSTM(X)}) ) 
\]
where, $X$ is the seuqence of input word embeddings and 
$f^{\text{pool}}$ is the pooling function to obtain an embedding for the sentence. 
They also concatenate the BiLSTM output with the GCN output.
\[
f_{enc}(X) = \text{FFNN}(\mathbf{h}_{\text{GCN}}(X); \mathbf{h}_{\text{BiLSTM}}(X))
\]
where FFNN is a feed-forward network.

The output from the encoder is fed to a classifier layer for cause and effect extraction. 
This model is trained to predict IOBES style tags for each word in the input.

\subsubsection{BERT Sequence Tagger}
\begin{figure*}[h]
    \centering
    \includegraphics[width=\linewidth]{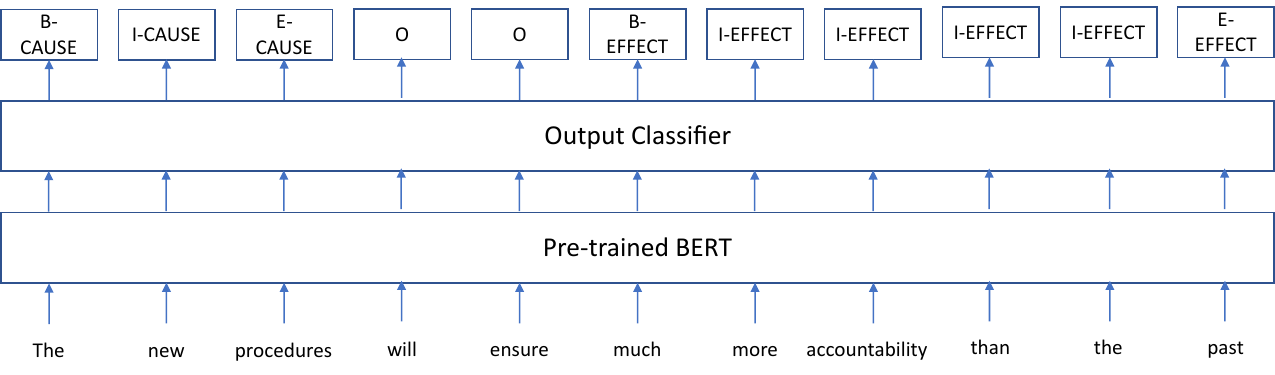}
    \caption{BERT Sequence Tagger}
    \label{fig:bert}
\end{figure*}

\paragraph{BERT-base} This model also uses the IOBES tagging scheme. 
We use the pretrained BERT-base \citep{devlin2018bert} model to obtain embeddings for the sentence. 
A softmax layer converts the output embeddings to label predictions. 
\[
y_i = \text{Softmax}(W . e_i + b)
\]
where, $e_i$ is the BERT embedding for the i-th token and $y_i$ is the softmax output. \\
This model (Fig. \ref{fig:bert}) is fine-tuned on the annotated data set to extract cause and effect spans.

\subsection{Span Based Model}
\begin{figure*}[h]
    \centering
    \includegraphics[width=\linewidth]{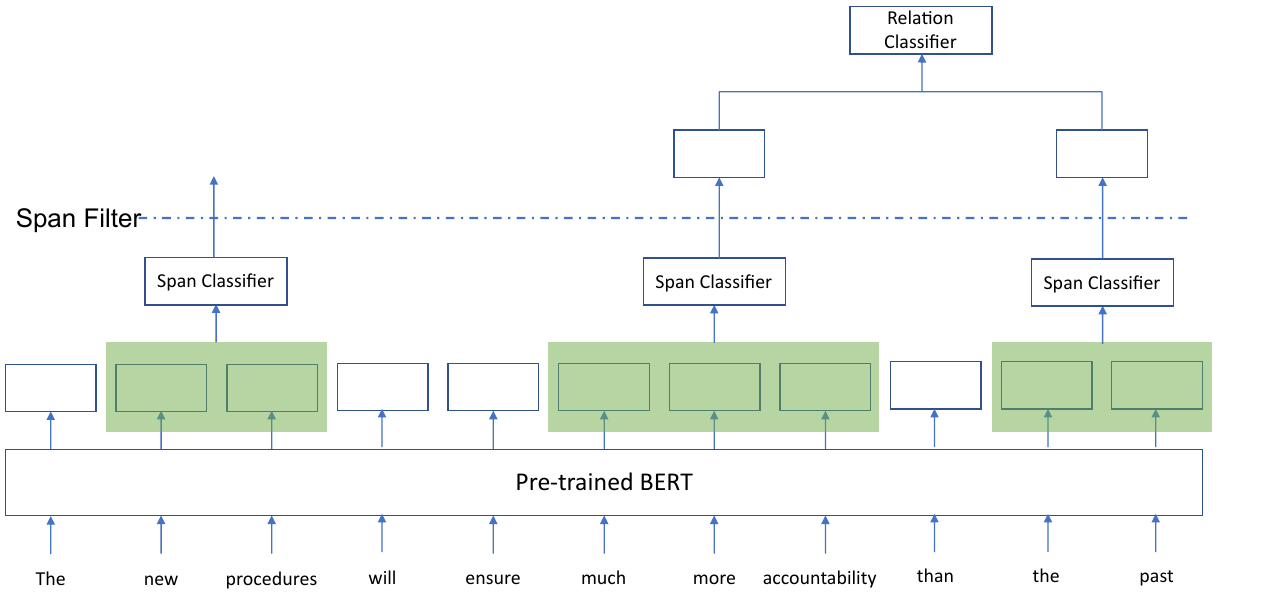}
    \caption{Causal SpERT Model}
    \label{fig:spert}
\end{figure*}

\paragraph{SpERT} In this model \citep{eberts2019span}, the spans of causes and effects are labeled along with the related pairs.
The model uses the output embeddings from BERT to classify candidate spans and their relations. (Fig. \ref{fig:spert}).
The model classifies a list of candidate spans generated by taking all possible sequences of words upto a maximum length from the sentence. 
The span representation is formed by combining the BERT embeddings of all tokens in the sentence.
A width embedding matrix is also trained to represent the size of the candidate span with a fixed size embedding. 
Span embedding is the concatenation of the max-pooled token embeddings in the span, the span size embedding and the [CLS] embedding from BERT.
\[
\mathbf{e}(s) = f(\mathbf{e}_i, \mathbf{e}_{i+1}, \dots \mathbf{e}_{i+k}) \circ w_{k+1} \circ c
\]
,where $\mathbf{e}(s)$ is the span embedding, $\mathbf{e}_i$ the embedding for i-th token and $w$ is the width embedding, $c$ is the CLS token embedding.
A span classifier labels the candidate spans as cause or effect or other using span embeddings.
\[
y_s = \text{softmax}(W_s . \mathbf{e}(s) + b_s)
\]
Then a relation classifier layer predicts the related pairs of spans out of all possible pairs.
A pair of spans is represented by the concatenation of the output embeddings from the span classifier and the max-pooled embeddings of the tokens in between the spans.

\section{Data Sets}

We use 4 causality extraction data sets in our experiments.
These data sets are converted into a common format (CREST from \cite{hosseini2021predicting}) to ensure the same labeling and evaluation method. 
There is one cause and one effect span in each example for all data sets.
To train the models, these data sets are preprocessed to different annotation templates from this common format.

% \paragraph{SemEval-2010} 
% This data set is from the SemEval 2010 competition task 8. 
% It was used for classification of semantic relation between pairs of noun phrases. 
% The entities are semantic heads of the nominals. 
% So most of the marked entities are single word. 
% There are \underline{1331} causal examples in the data set of \underline{10717} examples in total.

\begin{table}[h]
    \centering
    \begin{tabular}{|l|c|c|c|} \hline
    Data Set & Train Size & Dev Size & Test Size  \\ \hline
    SCITE & 4450 & 0 & 786 \\
    MedCaus & 8881 & 2963 & 2959 \\
    BeCauSE & 1226 & 0 & 354 \\
    FinCausal & 1044 & 347 & 348 \\ \hline
    \end{tabular}
    \caption{Data set statistics}
    \label{tab:data}
\end{table}

\begin{figure*}[h]
    \centering
    % \begin{subfigure}{.49\textwidth}
    %     \centering
    %     \includegraphics[width=\linewidth]{fig/semeval2010.png}
    %     \caption{Distribution of span length (SemEval-2010)}
    %     \label{fig:semeval2010}
    % \end{subfigure}
    \begin{subfigure}{.49\textwidth}
        \centering
        \includegraphics[width=\linewidth]{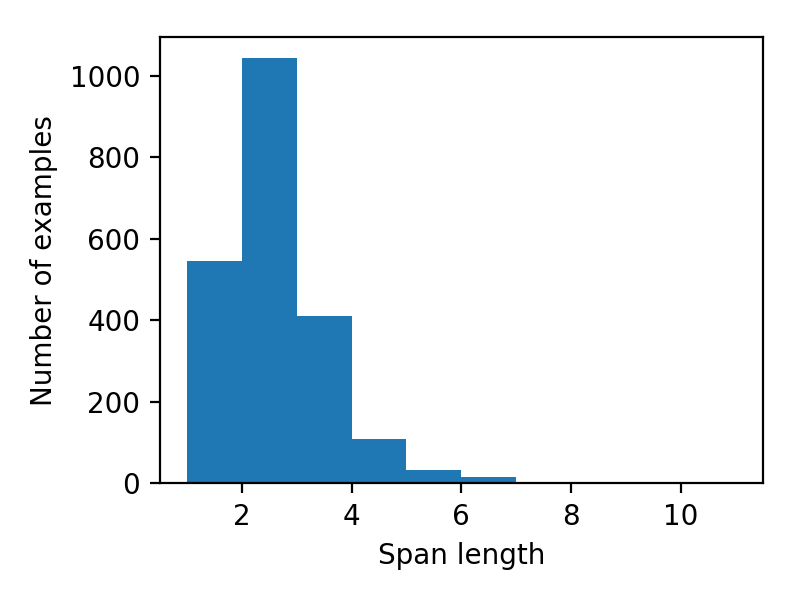}
        \caption{Distribution of span length (SCITE)}
        \label{fig:scite}    
    \end{subfigure}
    \begin{subfigure}{.49\textwidth}
        \centering
        \includegraphics[width=\linewidth]{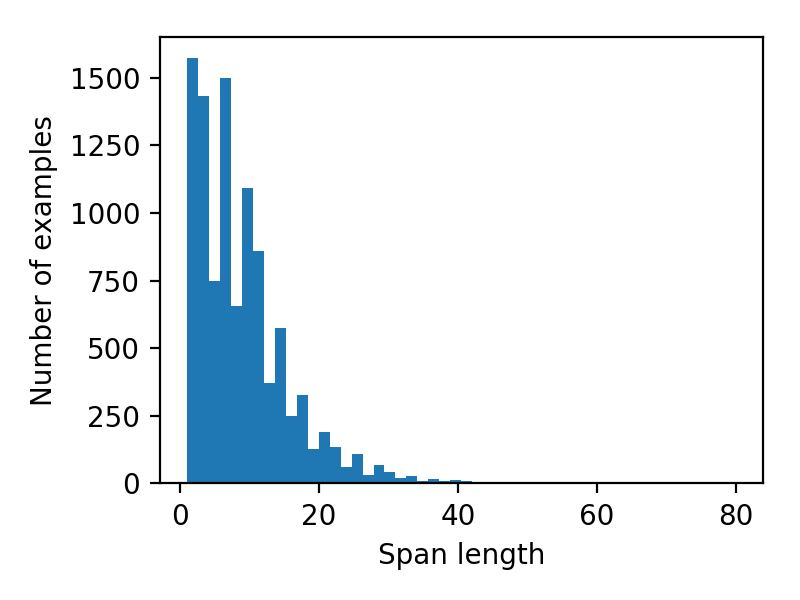}
        \caption{Distribution of span length (MedCaus)}
        \label{fig:medcaus}
    \end{subfigure}
    \begin{subfigure}{.49\textwidth}
        \centering 
        \includegraphics[width=\linewidth]{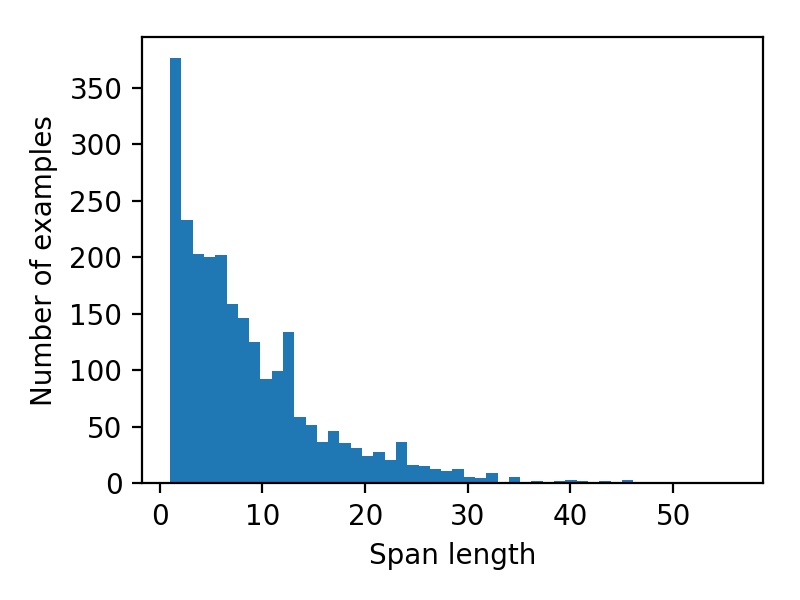}
        \caption{Distribution of span length (BeCauSE)}
        \label{fig:because}
    \end{subfigure}
    \begin{subfigure}{.49\textwidth}
        \centering 
        \includegraphics[width=\linewidth]{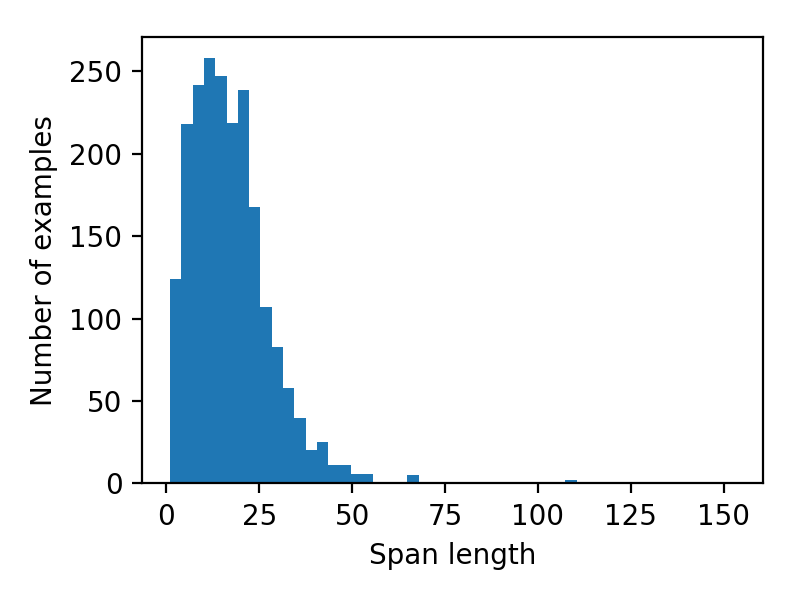}
        \caption{Distribution of span length (FinCausal)}
        \label{fig:fincausal}
    \end{subfigure}
    \caption{Distribution of span length for different data sets}
    \label{fig:span}
\end{figure*}

\paragraph{SemEval-2010 (SCITE)}
The SemEval-2010 data set was annotated by \cite{li2021causality} to mark the noun phrases instead of single words. 
They added tags for multiple cause-effect pairs in a sentence.
A new tag - Embedded Causality - was introduced to denote a cause or effect entity that is shared between multiple pairs.
The train and test sets have 4450 and 786 examples respectively.

\paragraph{MedCaus}
This is a medical causality data set with 15000 sentences used in \cite{moghimifar2020domain}. 
The examples were randomly extracted from medical articles in Wikipedia using manually selected causal cue words.
The cause and effect entities are long phrases in this data set.
There are 12064 causal examples and 3936 non-causal examples. 
The actual train, dev and test split used in the paper were not published. 
We randomly selected a 60-20-20 split of the whole data set after filtering examples with incorrect span labels.

\paragraph{BeCauSE}
The sentences for this data set \citep{dunietz2017because} were collected from multiples sources: New York Times, Wall Street Journal, Penn Tree Bank and Congress hearings.
The entities are also long phrases in this data set. 
There are 2150 examples in this data set.
We filter duplicate examples containing the same sentence and instances with wrong span labels.
After the preprocessing steps, we have \underline{1570} total samples.

\paragraph{FinCausal}
This is a financial domain data set \footnote{\url{https://github.com/yseop/YseopLab/tree/develop/FNP_2020_FinCausal/data}} with 1750 examples. 
After discarding incorrect labels there are \underline{1739} samples.
The cause and effect spans in this data set are whole sentences. So the contexts in this data set are paragraphs with multiple sentences.
The train, dev and test sets are selected randomly in a 60-20-20 split.

\section{Experiments}
% \subsection{SemEval-2010}

% \begin{tabular}{l|c|c|c} \hline
%     Model & Precision & Recall & F1 \\ \hline
%     BERT & 73.11 & 77.53 & 75.26 \\
%     SpERT (n=5) & \textbf{74.91} & \textbf{79.63} & \textbf{77.20} \\
%     SpERT (n=10) & 74.01 & 79.33 & 76.58 \\ \hline
% \end{tabular}

\subsection{SCITE}
We use the implementation of the model from \citet{li2021causality} \footnote{\url{https://github.com/Das-Boot/scite}} to run our experiments.
The hyperparameters are set to the values reported in the paper.
The batch size is 16 and the model is trained for 100 epochs for all data sets.

\subsection{GCE}
The dependency parsing for this model is obtained using the \texttt{spacy} library.
We train the GCE model with code from \citet{moghimifar2020domain}. \footnote{\url{https://github.com/farhadmfar/ace}}
We set the batch size to 20 and trained the model on causality extraction task for 50 epochs.
All other hyperparameters are set to the values used in \citet{moghimifar2020domain}.

\subsection{BERT}
The BERT sequence tagging model is implemented using the \textit{transformers} \cite{wolf-etal-2020-transformers} libray with PyTorch.
We set the batch size to 16 and train the model for 40 epochs with cross entropy loss.

\subsection{SpERT}
The SpERT model is trained with the implementation from \citet{eberts2019span}. 
% \footnote{\url{https://github.com/lavis-nlp/spert}}
We use the cause and effect spans as entities in the NER task.
The relation classifier predicts if a causal relation exists in the input text.
We train this model for 40 epochs with a batch size of 16.
The number of negative entities is set to 10 and negative relations to 5 to create training samples.
The model lists all possible spans of length less than the maximum span size and classifies them using the span classifier.
For predicting the cause and effect span in a text segment, we select the longest predicted span for each type.
The maximum span size, $n$ is set to a value based on the distribution of span length in the data set (Fig \ref{fig:span}).
We found the 99 percentile span size from the training data distribution to work well across data sets (marked by * in the result tables).
The code for this model can be found at \url{https://github.com/aniksh/causal-spert}.

\begin{table}[h]
    \fontsize{7.5}{9}\selectfont
    \centering
    \begin{tabular}{|l|c|c|c|} \hline
        Model & Precision & Recall & F1 \\ \hline
        SCITE & 88.19 (90.62) & 89.65 (89.77) & 88.91 (90.19) \\ 
        GCE & 87.64 (91.61) & 83.51 (84.73) & 85.52 (88.04) \\
        BERT & 90.50 (93.14) & 89.79 (89.51) & 90.14 (91.29) \\
        SpERT (n=4) & 91.89 (93.81) & 89.01 (88.09) & 90.43 (90.86) \\
        SpERT (n=6)* & 92.35 (92.52) & 88.48 (86.16) & 90.37 (89.22) \\
        SpERT (n=8) & 91.87 (92.36) & 88.74 (88.96) & 90.28 (90.63) \\ \hline
        % SCITE & 83.33 & 85.81 & 84.55 \\
        % SCITE (Cause) & 89.99 & 89.98 & 89.98 \\
        % SCITE (Effect) & 92.72 & 90.21 & 91.44 \\
        % SpERT (n=4) & 94.01 & 88.31 & 91.07 \\
        % SpERT (n=5) & \textbf{96.82} & 92.21 & \textbf{94.46} \\
        % SpERT (n=6) & 95.95 & 92.21 & 94.04 \\
        % SpERT (n=10) & 95.09 & 92.21 & 93.63 \\ \hline
    \end{tabular}
    \caption{Precision, Recall and F1 score for causality extraction on \underline{SCITE} data set (Partial matching score in bracket). $n$ is the maximum span size for SpERT.
    * denotes the 99 percentile span size. }
    \label{tab:scite}
\end{table}

% \begin{tabular}{l|c|c|c} \hline
%     Model & Precision & Recall & F1 \\ \hline
%     BERT & 74.12 & 72.9 & 73.5 \\
%     SpERT (n=10) & 60.27 & 42.13 & 49.59 \\
%     SpERT (n=15) & 55.20 & 54.24 & 54.72 \\
%     SpERT (n=20) & 52.64 & 59.88 & 56.03 \\
%     SpERT (n=25) & 40.11 & 61.36 & 48.51 \\ \hline
% \end{tabular}

% \begin{tabular}{l|c|c|c} \hline
%     Model & Precision & Recall & F1 \\ \hline
%     GCE & 54.31 & 56.97 & 55.61 \\
%     BERT & 59.23 & 58.51 & 58.87 &  \\
%     SpERT (n=10) & \textbf{65.39} & 43.79 & 52.45 \\
%     SpERT (n=15) & 60.72 & 55.46 & 57.97 \\
%     SpERT (n=20) & 57.43 & 61.21 & \textbf{59.26} \\
%     SpERT (n=25) & 55.81 & 63 & 59.18 \\
%     SpERT (n=25) & 51.68 & \textbf{64.58} & 57.41 \\ \hline
% \end{tabular}

\begin{table}[h]
\fontsize{7.5}{10}\selectfont
\begin{tabular}{|l|c|c|c|} \hline
    Model & Precision & Recall & F1 \\ \hline
    SCITE & 59.00 (71.23) & 54.74 (66.87) & 56.79 (68.98) \\
    GCE & 54.31 (70.06) & 56.97 (69.31) & 55.61 (69.68) \\
    BERT & 61.12 (75.73) & 60.38 (72.92) & 60.75 (75.29) \\
    % SpERT (n=10) & \textbf{79.02} & 42.98 & 55.68 & \textbf{91.70} & 31.15 & 46.51 \\
    SpERT (n=20) & 72.54 (88.80) & 57.96 (65.09) & 64.44 (75.12) \\
    SpERT (n=25) & 71.49 (86.54) & 59.92 (72.05) & 65.20 (78.63) \\
    SpERT (n=30) & 69.99 (85.90) & 60.21 (75.75) & 64.73 (80.51) \\
    SpERT (n=32)* & 71.52 (85.94) & 61.16 (76.12) & 65.93 (80.73) \\
    % SpERT (n=40) & 69.57 & 60.67 & 64.81 & 81.87 & 77.02 & 79.37 \\ 
     \hline
\end{tabular}
\caption{Precision, Recall and F1 score for causality extraction on \underline{MedCaus} data set (Partial matching score in bracket). $n$ is the maximum span size for SpERT.
* denotes the 99 percentile span size.}
\end{table}

\begin{table}[h]
    \centering
    \fontsize{7.5}{10}\selectfont
    \begin{tabular}{|l|c|c|c|} \hline
        Model & Precision & Recall & F1 \\ \hline
        SCITE & 37.74 (57.31) & 36.78 (54.08) & 37.25 (55.65) \\
        GCE & 27.15 (50.02) & 21.89 (53.23) & 24.24 (51.58) \\
        BERT & 51.46 (66.73) & 52.40 (68.29) & 51.92 (67.51) \\
        SpERT (n=20) & 53.92 (69.28) & 46.61 (50.96) & 50.00 (58.72) \\
        SpERT (n=25) & 54.01 (70.49) & 50.42 (61.63) & 52.15 (65.76) \\
        SpERT (n=30) & 54.68 (67.72) & 51.98 (64.53) & 53.29 (66.09) \\
        SpERT (n=34)* & 55.13 (67.60) & 52.40 (66.26) & 53.73 (66.92) \\ \hline
        % BERT & 79.15 & 73.25 & 76.09 \\
        % SpERT (n=10) & \textbf{88.17} & 64.19 & 74.29 \\
        % SpERT (n=15) & 84.96 & 79.22 & 81.99 \\
        % SpERT (n=20) & 83.28 & 87.50 & 85.34 \\
        % SpERT (n=25) & 86.82 & 91.22 & \textbf{88.96} \\
        % SpERT (n=30) & 81.20 & \textbf{93.41} & 86.88 \\ \hline
    \end{tabular}
    \caption{Precision, Recall and F1 score for causality extraction on \underline{BeCauSE} data set (Partial matching score in bracket). $n$ is the maximum span size for SpERT. * denotes the 99 percentile span size.}
    \label{tab:because}
\end{table}

\begin{table}[h]
\fontsize{7.5}{9}\selectfont
\centering
\begin{tabular}{|l|c|c|c|} \hline
    Model & Precision & Recall & F1 \\ \hline
    SCITE & 67.69 (79.41) & 68.59 (80.13) & 68.14 (79.77) \\
    GCE & 52.26 (69.04) & 51.14 (75.52) & 51.70 (72.14) \\
    BERT & 69.31 (77.24)  & 71.08 (81.70) & 70.18 (79.41) \\
    % SpERT (n=10) & 65.16 & 20.57 & 31.27 & 84.86 & 11.15 & 19.71 \\
    % SpERT (n=15) & 65.51 & 37.71 & 47.87 & 84.73 & 26.02 & 39.82 \\
    % SpERT (n=20) & 70.13 (77.28) & 47.29 () & 56.48 () &  & 36.89 & 49.94 \\
    % SpERT (n=25) & 50.91 & 63.86 & 56.65 & 76.35 & 56.85 & 65.17 \\ 
    % SpERT (n=30) & 73.23 (83.06) & 63.71 (64.43) & 68.14 (72.57) \\
    % SpERT (n=40) & 74.23 (81.92) & 68.71 (74.74) & 71.36 (78.16) \\
    % SpERT (n=50) & 73.38 (82.47) & 69.71 (81.56) & 71.50 (82.01) \\
    % SpERT (n=60) & 74.77 (83.79) & 70.71 (82.48) & 72.69 (83.13) \\ \hline
    SpERT (n=50) & 69.70 (80.96) & 66.71 (81.18) & 68.18 (81.07) \\
    SpERT (n=60) & 71.52 (82.39) & 67.43 (83.19) & 69.41 (82.79) \\
    SpERT (n=64)* & 70.61 (80.81) & 66.57 (83.81) & 68.53 (82.18) \\ \hline
    
    % SpERT (n=70) & 74.32 (79.86) & 69.86 (79.40) & 72.02 (79.63) \\ \hline
\end{tabular}
\caption{Precision, Recall and F1 score for causality extraction on \underline{FinCausal} data set (Partial matching score in bracket). $n$ is the maximum span size for SpERT. * denotes the 99 percentile span size. }
\end{table}

\section{Results}
Here we analyse the relation of the performance of our trained models to several factors.

\subsection{Evaluation}
The models are evaluated using micro F1 score for causality extraction.
For sequence tagging models, the predicted tags are converted to spans by detecting the start and end from the tags.
For IOBES tags, B and E denote the start and end.
With BIO tags, the end can be detected by finding the end of B or I tags.

\[
\text{precision} = \frac{\text{no. of correct predicted spans}}{\text{total predicted spans}}
\]

\[
\text{recall} = \frac{\text{no. of correct predicted spans}}{\text{total ground truth spans}}
\]

$$
\text{F1} = 2 * \frac{ \text{precision} * \text{recall} } { \text{precision} + \text{recall} }
$$

\begin{figure}[h]
    \centering
    \includegraphics[width=\linewidth]{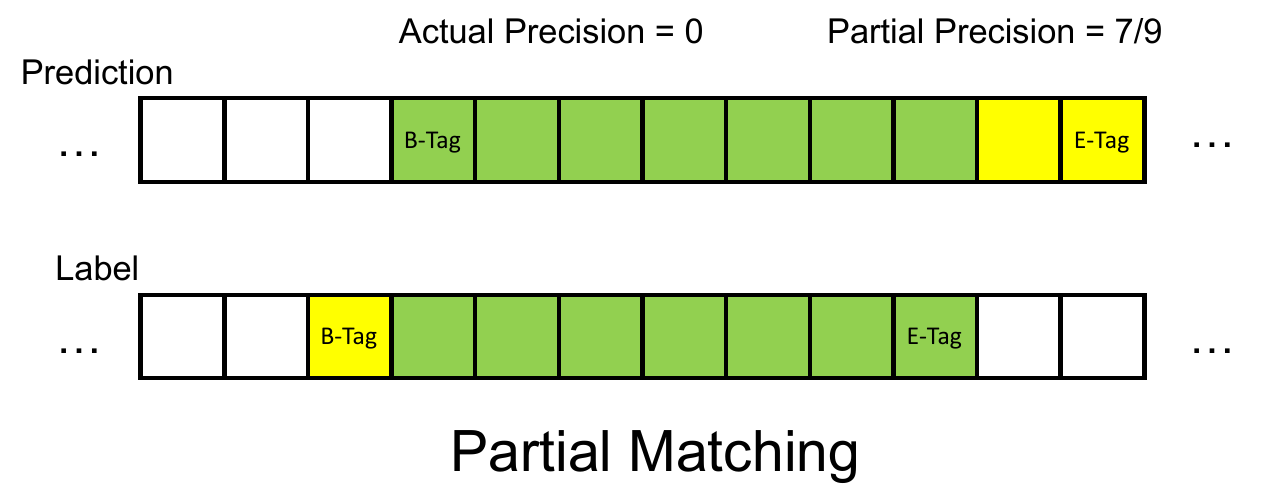}
    \caption{Exact vs. partial match}
    \label{fig:match}
\end{figure}

\paragraph{Exact vs. Partial Match}
As there are multiple words in a cause or effect span, it is possible that the predicted span will match with some of the words in the ground truth span. 
We can stipulate that the predicted span must exactly match the ground truth span, i.e. the start and end of the spans must be same.
We term this method as "Exact Match".
In "Partial Match", we count the number of common words between the predicted span and the ground truth span (Fig. \ref{fig:match}). 
So, the model is not penalized for missing one or two words from the ground truth span.

All models have a better F1 score while using partial matching.
The major difference in the predictions from sequence tagging and span based models is that the predicted spans cannot overlap in a sequence tagging model.
With the span based model, we need to select a span from possibly overlapping predicted spans.
In our experiments, we select the longest predicted span for each type.

\subsection{Data Set Variation}
\paragraph{Length of Spans}
The lengths of cause and effect spans in the SCITE data set are mostly below 5 with very few spans longer than that (Fig. \ref{fig:scite}).
The difference in the performance of the BERT and SpERT models is very low - 0.29\% for exact matching and 0.53\% for partial matching - in this data set. 
On the other data sets with longer span lengths, we see that the SpERT model outperforms all sequence tagging models in terms of exact match.
BERT has higher partial F1 score than BERT only on the BeCauSE data set.
As partial matching score depends on the length of the span, longer spans can skew this score even though SpERT performs better in terms of exact matching.
As the SpERT model predicts a contiguous span of words, it does not misclassify a span by missing 1 or 2 words in a long span.
So the span based model has a clear advantage in data sets with longer span lengths.

\paragraph{Presence of Causal Connective Word}
We can check the data sets for the presence of common causal connective words - because, due to, lead to - for estimating the percentage of explicitly causal sentences.
The MedCaus data set has the highest percentage of explicitly causal sentences with common causal indicators - $51\%$.
FinCausal has the lowest percentage at $8\%$ but the causes and effects in this data set are also sentences instead of phrases or clauses.
BeCauSE and SCITE data sets have similar percentage of explicit causality at $15\%$ each. 
We see that BERT and SpERT models get close score on both these data sets while there is a large gap in the MedCaus data set.
The GCE model also performs the best in the MedCaus data set.
So, the presence of causality indicators help the SpERT and GCE models.

\paragraph{Average Word Frequency of Spans}
We use Wikipedia word frequencies to get the average word frequency of a cause or effect span.
The FinCausal data set has the lowest word frequency as this is from the financial domain. 
MedCaus and BeCauSE data sets are similar in this metric and SCITE data set has the highest frequency.
So the spans in SCITE data set have the most common words. 
We see that there is very little difference in the performance of the model on this data set.
In the FinCausal data set, the SCITE model does much better than GCE as it uses different character-based representations of words.
It is clear that it is better to use data sets from different domains to evaluate the performance of these causality extraction models.

\subsection{Model}
The SpERT model has the best F1 score in terms of exact matching across all data sets. 
The advantage of this model is that we can select the maximum span size based on the distribution of the span lengths in the data set.
% BERT sequence tagging model outperforms SpERT in terms of partial match on the BeCauSE data set.

\paragraph{Parsing Information}
GCE model uses dependency parsing information to build the graph adjacency matrix used in the graph convolution layer (GCN) of this model.
But this model does not perform as well as the BiLSTM-CRF (SCITE) model which only takes the word sequence as input.
So the use of word dependency relation does not play a significant role in these experiments.

% \paragraph{Pre-trained Embeddings}
% Pre-trained embeddings play a vital role in the performance of neural models where training data is limited.
% The SCITE model performs better than GCE in data sets from different domains as it takes both word embeddings and character level embeddings - character CNN and Flair embeddings - to form word representation. 
% Both the BERT sequence tagger and SpERT are built on top of pre-trained BERT-base embeddings.
% So, they perform better than traditional word embedding based models in all data sets.

\paragraph{Pre-trained Language Model}
It is evident from the performance of BERT based models on 4 different data sets that pre-trained language models are much better than models only trained on the specific data set. 
A simple SoftMax classification layer on top of pre-trained BERT embeddings show significant improvement in F1 scores compared to both bi-directional LSTM and graph convolution encoder. 
As the data sets with labeled causes and effects have small number of training samples, it is important to augment the language comprehension ability of neural models using transfer learning methods.

\section{Conclusion}
In this work, we combine four data sets with cause and effect sequence labels into a standard format. 
We apply a span based approach to extracting causal knowledge that works better than the state-of-the-art across data sets from different domains.
This is the first work that applies span based models for causal knowledge extraction, with the following advantages over sequence tagging models: 1) We can tune the maximum span threshold for each domain/dataset, which enables the span-based model to consistently achieve the best performance across the domains and 2)
A span based model can handle multiple cause and effect entities.

% We show the relation between the data set attributes like span size distribution and average word frequency to modeling factors like pre-trained word embeddings.
We examine the relation of modeling methods like pre-training and additional parsing layer, and data set attributes like span size distributions to the causality extraction performance.
The results show the advantage of the span based approach to build neural models for data sets across diverse domains.

\bibliography{other}
\bibliographystyle{acl_natbib}

% \appendix

% \section{Example Appendix}
% \label{sec:appendix}

% This is an appendix.

\end{document}